\documentclass[]{fairmeta}

\usepackage{amsmath}
\usepackage{amssymb}
\usepackage{mathtools}
\usepackage{amsthm}
\usepackage{amsfonts}
\usepackage{algorithm}
\usepackage{algorithmic}
\usepackage{relsize}
\usepackage{nicefrac}
\usepackage{colortbl}
\usepackage{makecell}
\usepackage{longtable}
\usepackage{array}
\usepackage{tabularx}
\usepackage{etoc}
\usepackage[dvipsnames]{xcolor}
\usepackage{wrapfig}

\usepackage{listings}
\usepackage{etoc}
\theoremstyle{plain}

\theoremstyle{definition}

\theoremstyle{remark}

\title{Mid-Training Language Models on Raw Video}

\author[1,2,*]{Jaedong Hwang}
\author[1,3,*]{Xiaoqian Shen}
\author[1]{Ernie Chang}
\author[1]{Changsheng Zhao}
\author[1]{Chong Zhou}
\author[1]{Saksham Suri}
\author[1]{Qi Qian}
\author[1]{Zechun Liu}
\author[1]{Lemeng Wu}
\author[1,4,*]{Qinsi Wang}
\author[1]{Raghuraman Krishnamoorthi}
\author[1]{Wei Wen}

\affiliation[1]{Meta AI}
\affiliation[2]{Massachusetts Institute of Technology}
\affiliation[3]{KAUST}
\affiliation[4]{Duke University}

\contribution[*]{Work done at Meta}

\newcommand{\ie}{\textit{i.e.}}

\abstract{

Multimodal large language models learn mostly from paired image-text data or annotated video, and raw web video is rarely used to further train an existing language model.
We study whether raw video, with no captions and no text loss, can serve as mid-training data for a pretrained language model.
Frames are encoded into continuous visual tokens, and the language model learns to predict the next visual token.
We mid-train Qwen3-1.7B on raw clips from YT-Temporal-1B and then apply the same image-text instruction tuning to it and to the model without mid-training, so that the two differ only in mid-training.
The mid-trained model scores 2.9 points higher on average across four video benchmarks and 5.1 points higher across ten image benchmarks, spanning perception, document, and chart tasks. 
Text performance is preserved even though mid-training includes no text, with an average of 48.9 across 14 text benchmarks compared with 48.0 for the model without mid-training.
Analyses across training show that the image and video gains emerge within 30\% of training and plateau thereafter, varying by less than 0.5 points.
Predicting captions fails to outperform next-visual-token prediction, demonstrating that video mid-training can remain purely self-supervised without the computational overhead or labeling noise of automated captioning.}

\date{\today}
\correspondence{Wei Wen at \email{wewen@meta.com}}

\metadata[Project Page]{\url{https://jd730.github.io/projects/RawVideoMidTraining}}

\begin{document}

\let\cite\citep

\maketitle


\section{Introduction}
\label{sec:intro}

Large language models (LLMs) have advanced rapidly across reasoning, coding, and knowledge-intensive tasks~\citep{achiam2023gpt,brown2020language,team2025gemma,yang2025qwen3}.
These capabilities have been driven in large part by data scaling, with recent models trained on tens of trillions of tokens, well beyond compute-optimal allocations~\citep{hoffmann2022training,yang2025qwen3}.
However, training set growth already outpaces human text generation, and the stock of public human text is projected to be fully used in a decade~\citep{villalobos2024will}.
Scaling the text pretraining corpus, a central driver of progress under neural scaling laws~\citep{kaplan2020scaling,hoffmann2022training}, is therefore approaching fundamental limits.
Repeated text yields diminishing returns beyond a few epochs~\citep{muennighoff2023scaling}, and recursive training on synthetic text degrades model quality~\citep{shumailov2024ai}.
Sustaining scaling trends necessitates drawing upon non-text modalities.

Among non-text modalities, video is the largest source of data.
Its stock is estimated at roughly 1350 trillion text-token equivalents, about 2.6 times the indexed web~\citep{villalobos2024will}.
Video also suits language model training, since its frames arrive in temporal order like tokens in text and show objects, actions, and scenes that text describes only indirectly.
Yet current multimodal models use video mainly through text, pairing sampled frames with captions or question-answer annotations~\citep{an2026llava,bai2025qwen3}.
Captioning web-scale video frame by frame is expensive, and the captions inherit the errors and omissions of the captioning model.
This raises the question of whether a pretrained language model can learn from raw video directly, with no captions, by predicting the next visual token.

While world models trained on large-scale video show that predicting future observations, in pixel space or in a learned representation space, supports simulation, planning, and control~\citep{assran2025v,bruce2024genie,agarwal2026cosmos}, integrating raw video into language model pretraining remains challenging.
Prior multimodal efforts either train unified transformers from scratch on text, image, and video tokens~\citep{wang2024emu3} or add latent future-frame prediction heads over curated video clips~\citep{yang2026cambrian}.
Neither directly addresses the most practical regime for scaling, namely continued pretraining of an existing, pretrained LLM on raw web video.

In this work, we mid-train a pretrained language model on raw video with next-token prediction over visual tokens.
Each frame is encoded by a pretrained vision encoder and projected into the language model's embedding space, turning a clip into a continuous sequence of visual tokens.
The language model learns to predict each next visual token from its hidden state, with no captions or discrete visual tokenizer.

We train Qwen3-1.7B~\citep{yang2025qwen3} this way on 16-frame clips from YT-Temporal-1B~\citep{zellers2022merlotreserve} and compare it with a baseline that receives the same instruction tuning without mid-training.
Video mid-training raises the average over four video benchmarks by 2.9 points, with a 4.8-point gain on EgoSchema, and the average over ten image benchmarks by 5.1 points.
Although mid-training includes no text, the mid-trained model preserves text performance, raising the average text score from 48.0 to 48.9 and keeping every knowledge and commonsense score within 1.4 points of the model without mid-training.
Replacing the visual target with a caption of the next frame gives similar video and image scores but lowers the average text score by 1.5 points, and a caption of the current frame leaves the image average at 48.8, within 0.1 points of the baseline.
Video mid-training therefore needs no frame-level captions.

Our primary contributions are as follows:
\begin{itemize}
    \item We propose mid-training a pretrained language model on raw video by predicting the next visual token, which requires no captions or annotations.
    \item We show that video mid-training improves video understanding by 2.9 points on average (4.8 points on EgoSchema) and image understanding by 5.1 points, while maintaining baseline text performance despite using no text supervision.
    \item We demonstrate that video mid-training eliminates the need for expensive captioning pipelines, as predicting current- or next-frame captions provides no advantage over purely self-supervised next-visual-token prediction.
\end{itemize}


\section{Related Work}
\label{sec:related}

\subsection{Learning from Video without Language}
Self-supervised video models are trained by reconstructing masked spatiotemporal patches~\citep{tong2022videomae}, autoregressively generating discrete visual tokens~\citep{yan2021videogpt}, or predicting representations in a latent feature space~\citep{assran2023ijepa,assran2025v}.
When scaled to internet video, predictive objectives yield rich representations that support physical planning~\citep{assran2025v} as well as generative world models that simulate interactive environments~\citep{bruce2024genie,agarwal2026cosmos}.
However, these models are trained strictly from scratch in the visual domain and typically incorporate language only post hoc through downstream probing or fine-tuning.
In contrast, rather than training a standalone video backbone, our objective is to investigate what an already-competent language model acquires when trained directly on continuous visual streams.

\subsection{Video as Training Data for Language Models}
An emerging line of research explores video as pretraining data for language architectures.
\citet{tong2026beyond} pretrain a transformer from scratch on a joint mixture of text and raw video, finding that adding video keeps text perplexity close to text-only pretraining, lower on in-distribution web text but higher on an out-of-distribution corpus.
However, their setup trains jointly from scratch with an explicit text language modeling loss.
Similarly, Emu3~\citep{wang2024emu3} pretrains a single transformer from scratch across interleaved discrete image, video, and text tokens.
Cambrian-S~\citep{yang2026cambrian} adds a latent frame prediction head to a vision-language model and trains it jointly with spatial instruction tuning on a curated video corpus.
Other recent efforts cast diverse multimodal tasks as video frame prediction~\citep{everythingvideo} or train vision-language models to anticipate and describe future environmental states~\citep{vlwm}.
Unlike these methods, we continue training an existing pretrained language model instead of training from scratch, and we use raw web video with no captions and no text loss.
The model learns only to predict the next continuous visual token, so video extends its autoregressive pretraining rather than serving as a task format or an instruction-tuning signal.

\subsection{Vision-Language Models and Mid-Training Data}
Standard vision-language models connect a pretrained vision encoder~\citep{zhai2023sigmoid} to a causal language model~\citep{yang2025qwen3} via a learned projection layer, followed by supervised instruction tuning on multimodal data~\citep{tong2024cambrian,li2024llavaonevision}.
Under this conventional recipe, the language model encounters visual tokens primarily during instruction tuning, and video is reduced to a sparse subset of sampled frames.
To bridge the gap between static image pretraining and downstream tasks, recent open models add a large-scale multimodal mid-training stage in which video is supervised mainly through synthesized clip captions~\citep{an2026llava,bai2025qwen3}.
Our ablations show that next-frame captions give no gain over predicting continuous visual features directly (see Section~\ref{subsec:objective}).
Bypassing caption generation lets video mid-training use large web collections such as YT-Temporal-1B~\citep{zellers2022merlotreserve} without any annotation.


\section{Next-Token Prediction on Visual Tokens}
\label{sec:method}

\begin{figure}[t]
\centering
\includegraphics[width=\linewidth]{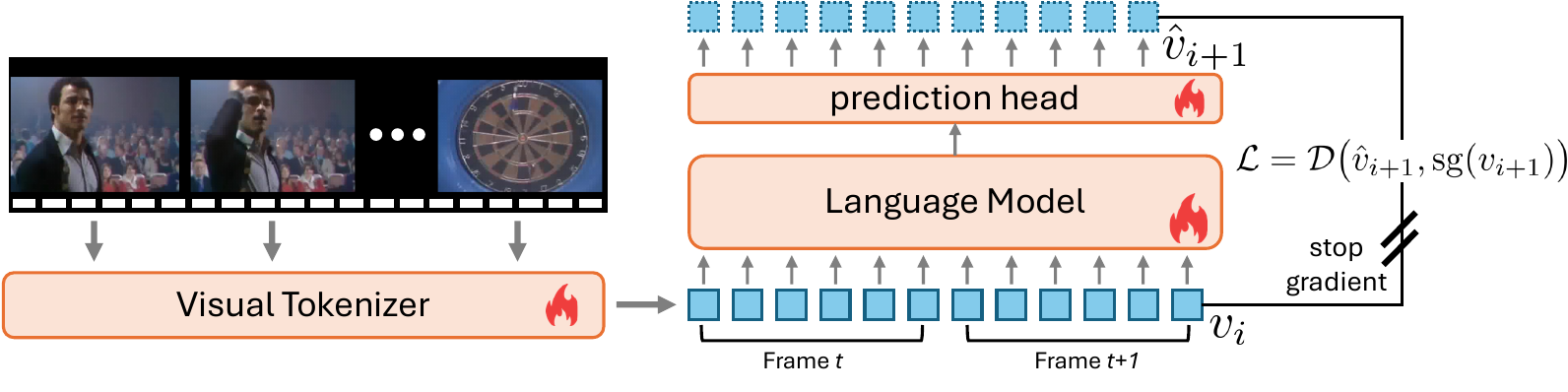}
\vspace{-0.6cm}
\caption{\textbf{Overview of continuous video mid-training.}
Ordered video frames $x_{1:T}$ are encoded and projected into continuous token representations.
Spatial patch grids are serialized with row-delimiter tokens and concatenated into a unified sequence $V$.
A causal language model processes the continuous visual stream without captions or text supervision, and a prediction head $r_\omega$ regresses the subsequent visual representation $v_{i+1}$ from the hidden state $h_i$.
A stop-gradient operator $\operatorname{sg}(\cdot)$ is applied to the regression target so that gradients do not flow through it.
}
\label{fig:method}
\end{figure}

We formulate video mid-training as autoregressive next-token prediction over continuous visual representations, keeping the causal decoder-only architecture of the language model.
Rather than quantizing video into discrete codebooks~\citep{yan2021videogpt} or relying on generated text captions~\citep{an2026llava}, our approach encodes raw video frames into continuous feature vectors that directly populate the input space of the language model backbone.
The model is trained on this serialized visual stream via self-supervised feature regression, without paired captions or any language loss (see Figure~\ref{fig:method}).
This enables seamless data scaling by bypassing the cost and complexity of manual annotation or synthetic pseudo-captioning pipelines.

\paragraph{Architecture.}
Given an ordered sequence of $T$ video frames $x_{1:T} = [x_1, \dots, x_T]$, a vision encoder $g_\phi$ and projector $p_\psi$ map each frame $x_t$ into the language model's hidden dimension $d$:
\begin{equation}
z_t = p_\psi(g_\phi(x_t)) \in \mathbb{R}^{N \times d},
\end{equation}
where $N$ denotes the number of visual tokens per frame.
Each grid $z_t$ is serialized in row-major order with a learned newline embedding $e_{\mathrm{nl}} \in \mathbb{R}^{d}$ after every row, which keeps the two-dimensional layout recoverable from the flat sequence.
The per-frame sequences are temporally concatenated into a unified continuous token stream $V = [v_1, \dots, v_L] \in \mathbb{R}^{L \times d}$, where $L$ is the total sequence length including the newline embeddings.
These embeddings are directly fed to the causal Transformer blocks of $f_\theta$ in the language model.

\paragraph{Objective.}
Standard language model pretraining maximizes the log-likelihood of discrete token sequences under causal masking.
For continuous visual representations where no discrete vocabulary exists, we frame next-token prediction as self-supervised feature regression in the embedding space (\ie, token space), similar to V-JEPA 2~\citep{assran2025v}, which regresses the embeddings of masked video regions.
At each position $i \in \{1, \dots, L-1\}$, the language model produces a contextual hidden state $h_i = f_\theta(v_{1:i})$ conditioned strictly on the preceding visual context.
A two-layer MLP prediction head $r_\omega: \mathbb{R}^d \to \mathbb{R}^d$ maps $h_i$ to a predicted feature vector $\hat{v}_{i+1} = r_\omega(h_i)$.
The model minimizes the distance between the predicted representation and the ground-truth visual token $v_{i+1}$ at every position whose next token is a visual patch,
\begin{equation}
    \mathcal{L} = \frac{1}{|L-1|}\sum_{i = 1}^{L-1} \mathcal{D}\big(\hat{v}_{i+1}, \operatorname{sg}(v_{i+1})\big),
    \label{eq:serial-loss}
\end{equation}
where $\mathcal{D}(u, v)$ and $\operatorname{sg}(\cdot)$ denote a distance function and the stop-gradient operator.
In this work, we use cosine distance $\mathcal{D}(u, v)= 1 - \frac{u^\top v}{\|u\|_2 \|v\|_2}$.
Applying $\operatorname{sg}(\cdot)$ to the target representation $v_{i+1}$ keeps gradients from flowing through the target.
This ensures that gradients propagate solely through the causal predictive pathway from the vision encoder to the prediction head ($\phi, \psi, \theta, \omega$), while the subsequent visual feature acts as a fixed regression target during each optimization step.
Under standard causal self-attention over the serialized sequence, this single objective unifies spatial next-patch modeling within frames and temporal prediction across frame boundaries.
It adds only a small prediction head used during mid-training and requires no discrete codebook or language supervision.


\section{Experiments}
\label{sec:experiments}

We first outline the experimental setup, then report downstream evaluations across video, image, and text benchmarks in Tables~\ref{tab:main-video} to~\ref{tab:main-text}, and finally analyze cross-modal training dynamics and the mid-training objective through ablation experiments.

\subsection{Experimental Details}
\label{subsec:exp_details}

\subsubsection{Architecture}
Following Cambrian-S~\citep{yang2026cambrian}, our model connects a pretrained vision encoder to a causal language model through a two-layer GELU~\citep{hendrycks2016gaussian} MLP projector.
 We instantiate the visual backbone with SigLIP SO400M/14 at $384\times384$ resolution~\citep{zhai2023sigmoid} and the language model with Qwen3-1.7B~\citep{yang2025qwen3}.
Visual patch tokens are projected directly into the 2,048-dimensional language embedding space.
The auxiliary prediction head $r_\omega$ used during mid-training is a lightweight two-layer GELU MLP with 2,048 hidden units.

\subsubsection{Dataset}
We draw mid-training video from YT-Temporal-1B (YT-1B)~\citep{zellers2022merlotreserve}, a collection of approximately 20 million public YouTube videos.
After excluding videos that are no longer accessible online, our source pool comprises 17 million videos.
Frames are sampled at 0.2\,fps (one frame every five seconds).
From videos containing between 16 and 63 sampled frames, we extract contiguous 16-frame clips starting at random temporal offsets.
The composed dataset contains 6.99M clips with 64.4B visual tokens in our implementation.
For subsequent supervised fine-tuning, we incorporate LLaVA-OneVision-Data~\citep{li2024llavaonevision} (3.9M images, 3.2B tokens).

\subsubsection{Captioning Mid-Training Data for Baselines}
\label{subsubsec:captioning}
For the caption baselines in Section~\ref{subsec:objective}, frames are captioned by Qwen3-VL-30B-A3B~\citep{bai2025qwen3} in consecutive pairs with a progress-aware prompt adapted from \citet{xue2025progress}.
The captioning model sees both frames of a pair together and writes one caption per frame that describes the action in that frame without referring to the other frame, so the two captions differ where the action has progressed.
The caption of frame $t$ is the target of current-caption prediction, and the caption of frame $t{+}1$ is the target of next-caption prediction from frame $t$.

The system prompt instructs the model to describe only what is visible and never to invent content absent from the frames.
The user turn presents the two frames in order and closes with the instruction reproduced in Figure~\ref{fig:caption-prompt}.

\begin{figure}[t!]
\centering
\begin{tikzpicture}
\node[draw=black!55, line width=0.6pt, rounded corners=3pt, inner sep=7pt]{%
\begin{minipage}{0.9\linewidth}
\scriptsize\ttfamily\raggedright
\textbf{System.} You are a precise video-frame captioner. You are shown an ordered sequence of keyframes from a single video. Describe what is VISIBLE in each frame -{}- people, objects, actions, setting, camera. Never invent things that are not in the frame.\\[5pt]
\textbf{User.} Frame 1: \textless image\textgreater{} Frame 2: \textless image\textgreater\\[2pt]
These 2 frames are consecutive moments of one action, in order. Write one caption per frame. Requirements:\\
(1) Ensure each frame's description is specific to the corresponding frame, not referencing other frames (no 'continues', 'now', 'still', 'again', 'then', 'previous', 'next').\\
(2) The description should focus on the specific action being performed, capturing the progression of the action. There is no need to comment on other elements, such as the background or unrelated objects.\\
- 1-3 concrete sentences; keep every caption distinct.\\
- Write a plain natural-language sentence for each frame. Do NOT output bounding boxes, coordinates, object labels, or detection JSON.\\
- Output ONLY a JSON array of exactly 2 plain strings (one per frame, in order), e.g. ["caption 1", "caption 2", ...] -{}- no objects, no extra text.
\end{minipage}};
\end{tikzpicture}
\caption{\textbf{Captioning prompt for the caption baselines.}
Qwen3-VL-30B-A3B receives two consecutive frames and writes one caption per frame, following a prompt adapted from the progress-aware captioning of \citet{xue2025progress}.
Each caption describes the action in its own frame without referring to the other frame, and each \textless image\textgreater{} is replaced by the visual tokens of that frame.}
\label{fig:caption-prompt}
\end{figure}

\subsubsection{Training}
All model parameters ($\phi, \psi, \theta, \omega$) are trained end-to-end for one epoch (27,300 steps) across 128 NVIDIA H100 GPUs.
We train with an effective batch size of 256 clips and a context window of 16k tokens.
We use AdamW~\citep{loshchilov2019decoupled} with a learning rate of 0.00001 for the language model and projector and 0.000002 for the vision encoder, with cosine decay and a 3\% linear warmup.
Following mid-training, we instruction-tune mid-trained models for one epoch on LLaVA-OneVision data~\citep{li2024llavaonevision} using the same training parameters used in mid-training.

\subsubsection{Benchmarks}
We evaluate instruction-tuned models across video, static image, and text suites.
\textbf{Video understanding.}
We group the video benchmarks by the ability they test, namely long-form egocentric understanding with EgoSchema~\citep{mangalam2023egoschema}, general video question answering with NExT-QA~\citep{xiao2021nextqa} and Video-MME~\citep{fu2024videomme}, and fine-grained
  temporal reasoning with TempCompass~\citep{liu2024tempcompass}.
\textbf{Image understanding.}
We group the image benchmarks by task type, namely text-rich image understanding~\citep{masry2022chartqa,mathew2021docvqa,mathew2022infographicvqa,singh2019textvqa}, general perception and diagram understanding~\citep{kembhavi2016ai2d,liu2024mmbench,li2024seedbench}, text recognition~\citep{liu2024ocrbench}, and visual reasoning~\citep{chen2024mmstar,lu2024mathvista}.
\textbf{Text understanding and reasoning.}
To measure how well the model retains its language ability, we evaluate commonsense reasoning and world knowledge~\citep{bisk2020piqa,clark2018think,clark2019boolq,hendrycks2021measuring,mihaylov2018can,sakaguchi2021winogrande,sap2019social,zellers2019hellaswag}, mathematical, scientific, and multi-step reasoning~\citep{cobbe2021gsm8k,hendrycks2021math,lightman2024let,rein2023gpqa,suzgun2023challenging}, and code generation~\citep{austin2021mbpp}.

\subsubsection{Evaluation Protocol}
Image and video benchmarks are evaluated via \texttt{lmms-eval}~\citep{zhang2024lmmsevalrealitycheckevaluation}, where multiple-choice questions are scored by option-letter log-likelihood and open-ended queries are decoded greedily.
Multiple-choice and short-answer text benchmarks are evaluated zero-shot by ranking candidate-answer log-likelihoods.
Free-form text benchmarks use greedy decoding with standard few-shot exemplars: eight for GSM8K, and three for MBPP and BBH.
Finally, GPQA is evaluated via \texttt{lighteval}~\citep{lighteval} in non-reasoning mode with an 8,192-token generation budget, a presence penalty of 1.5, and top-$k=20$.

\begin{table}[t]
\caption{
Effect of video mid-training on video benchmarks.
\textit{No Mid-Training} denotes the language model instruction-tuned without video mid-training, and \textit{Video Mid-Training} denotes the model mid-trained for one epoch (64.4B visual tokens) and then given the same instruction tuning.
All scores are accuracies in percent, and the best value in each row is in bold.
}
\vspace{0.1cm}
\label{tab:main-video}
\centering
\setlength{\tabcolsep}{3pt}
\resizebox{0.5\linewidth}{!}{%
\begin{tabular}{lcc}
\toprule
Benchmark & No Mid-Training & Video Mid-Training \\
\midrule
\multicolumn{3}{l}{\textit{Text-Rich Image Understanding}} \\
ChartQA        & 43.76 & \textbf{51.96} \\
DocVQA         & 32.13 & \textbf{38.57} \\
InfographicVQA & 22.31 & \textbf{25.28} \\
TextVQA        & 45.49 & \textbf{51.05} \\
\midrule
\multicolumn{3}{l}{\textit{General Perception \& Diagrams}} \\
AI2D           & 66.39 & \textbf{72.51} \\
MMBench        & 76.00 & \textbf{79.46} \\
SEED-Bench     & 63.00 & \textbf{69.10} \\
\midrule
\multicolumn{3}{l}{\textit{Character Recognition}} \\
OCRBench       & 38.40 & \textbf{41.60} \\
\midrule
\multicolumn{3}{l}{\textit{Complex Visual Reasoning}} \\
MathVista      & 58.15 & \textbf{62.59} \\
MMStar         & 41.60 & \textbf{45.93} \\
\midrule
Average        & 48.72 & \textbf{53.80} \\
\bottomrule
\end{tabular}
}
\end{table}

\begin{table}[t]
\caption{Effect of video mid-training on image benchmarks.
Columns follow Table~\ref{tab:main-video}.
All scores are accuracies in percent, with OCRBench divided by ten, and the best value in each row is in bold.
}
\vspace{0.1cm}
\label{tab:main-image}
\centering
\setlength{\tabcolsep}{3pt}
\resizebox{0.5\linewidth}{!}{%
\begin{tabular}{lcc}
\toprule
Benchmark & No Mid-Training & Video Mid-Training \\
\midrule
NExT-QA               & 59.04 & \textbf{62.18} \\
VideoMME              & 42.85 & \textbf{45.15} \\
TempCompass           & 51.58 & \textbf{53.04} \\
EgoSchema             & 45.20 & \textbf{50.00} \\
\midrule
Average               & 49.67 & \textbf{52.59} \\
\bottomrule
\end{tabular}
}
\end{table}

\begin{table}[t]
\caption{Effect of video mid-training on text benchmarks.
Qwen3-1.7B is the language model before any multimodal training.
All scores are accuracies in percent, and the best value in each row is in bold.
}
\vspace{0.1cm}
\label{tab:main-text}
\centering
\setlength{\tabcolsep}{3pt}
\resizebox{0.6\linewidth}{!}{%
\begin{tabular}{lccc}
\toprule
Benchmark & Qwen3-1.7B & No Mid-Training & Video Mid-Training \\\midrule
\multicolumn{4}{l}{\textit{Commonsense \& General Knowledge}} \\
OpenBookQA           & 25.99          & \textbf{26.79} & 26.59 \\
WinoGrande           & 60.53          & \textbf{61.87} & 61.64 \\
HellaSwag            & 45.18          & 45.26          & \textbf{45.68} \\
SIQA                 & 44.03          & 44.18          & \textbf{44.49} \\
PIQA                 & 71.47          & \textbf{72.34} & 72.17 \\
ARC-Easy             & 71.58          & 73.31          & \textbf{73.61} \\
ARC-Challenge        & 39.81          & 41.35          & \textbf{42.72} \\
BoolQ                & 79.46          & 80.59          & \textbf{81.54} \\
MMLU                 & \textbf{62.35} & 61.16          & 60.75 \\
\midrule
\multicolumn{4}{l}{\textit{Mathematical \& Complex Reasoning}} \\
MATH500              & \textbf{45.63} & 27.98          & 29.17 \\
GSM8K                & \textbf{73.41} & 49.92          & 54.77 \\
BBH                  & \textbf{38.89} & 23.92          & 29.40 \\
GPQA                 & 18.00          & \textbf{19.50} & 17.50 \\
\midrule
\multicolumn{4}{l}{\textit{Code Generation}} \\
MBPP                 & 43.06          & 44.25          & \textbf{44.44} \\
\midrule
Average              & \textbf{51.38} & 48.03          & 48.89 \\
\bottomrule
\end{tabular}
}
\end{table}

\begin{figure}[t]
\centering
\includegraphics[width=0.75\linewidth]{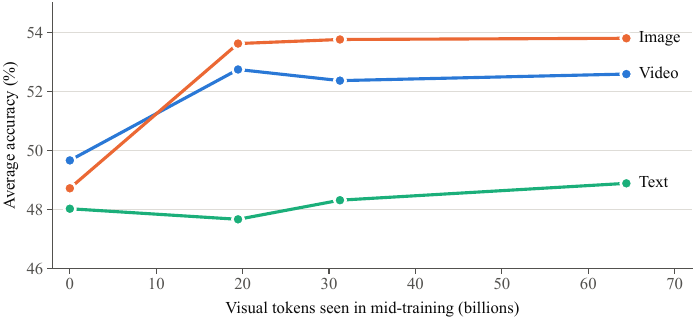}
\caption{\textbf{Progress over mid-training.}
Each point corresponds to an intermediate mid-training checkpoint followed by identical instruction tuning (0.3, 0.5, and 1.0 epochs), plotted against cumulative visual tokens processed during mid-training.
The point at zero represents the baseline instruction-tuned model without video mid-training.
Figure~\ref{fig:progress-all} reports individual benchmark trajectories.}
\label{fig:progress}
\end{figure}

\subsection{Video Mid-Training Improves Video and Image Understanding}
\label{subsec:main_results}

Tables~\ref{tab:main-video}, \ref{tab:main-image}, and~\ref{tab:main-text} report downstream evaluation results across video, image, and text benchmarks, respectively.
Compared with the model instruction-tuned without mid-training, our video mid-trained model improves every video benchmark (+2.9 points on average), with the largest gain of 4.8 points on EgoSchema.
These representations transfer to static image perception, improving all ten image benchmarks (+5.1 points on average), with the largest gains on visually dense tasks such as ChartQA (+8.2 points), DocVQA (+6.4 points), and AI2D (+6.1 points).

On text benchmarks, multimodal instruction tuning lowers performance on mathematical and multi-step reasoning, dropping 23.5 points on GSM8K, 17.7 points on MATH500, and 15.0 points on BBH relative to the original Qwen3-1.7B, while leaving knowledge and commonsense benchmarks largely unaffected.
Prior vision-language models report the same loss and reduce it by mixing text-only data into training~\citep{lu2024deepseek}, whereas our instruction-tuning set contains only image-text examples.
The mid-trained model shows the same pattern with a smaller drop, 18.6 points on GSM8K, 16.5 points on MATH500, and 9.5 points on BBH.
One possible reason is that mid-training has already adapted the language model to visual tokens, so instruction tuning needs to change it less.

\begin{figure}[t!]
\centering
\includegraphics[width=0.75\linewidth]{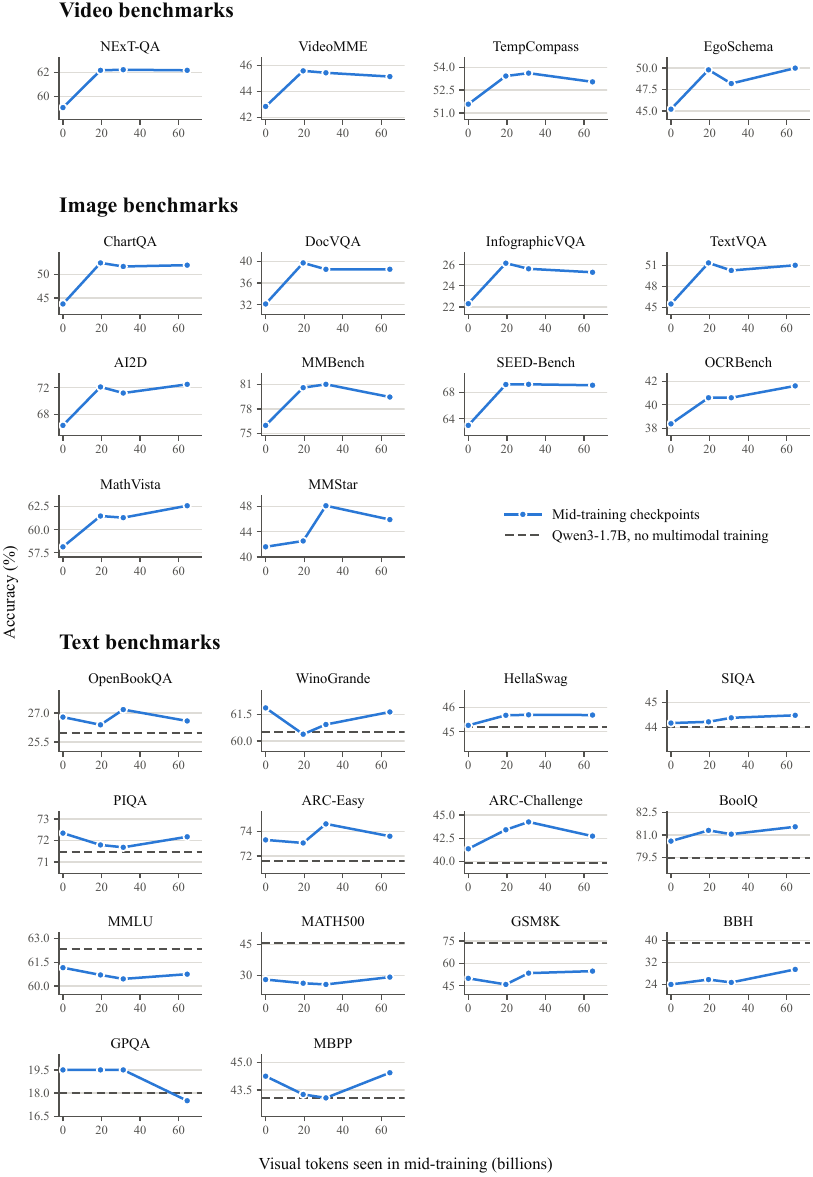}
\caption{
\textbf{Performance trajectory across mid-training on each benchmark.}
Each point corresponds to an intermediate mid-training checkpoint after subsequent instruction tuning, plotted against cumulative visual tokens processed during mid-training.
The point at zero represents the baseline model without mid-training.
In text benchmark panels, the dashed line indicates the original Qwen3-1.7B before multimodal training.
}
\label{fig:progress-all}
\end{figure}

\subsection{Image and Video Gains Appear Early in Mid-Training}
\label{subsec:progress}
Figure~\ref{fig:progress} illustrates average video, image, and text performance across mid-training, evaluated at 0.3, 0.5, and 1.0 epochs.
At each training stage, we perform supervised fine-tuning on LLaVA-OneVision-Data~\citep{li2024llavaonevision} to evaluate downstream multimodal capabilities.
Average image and video scores rise by 4.9 and 3.1 points within the first 0.3 epochs (19.5B visual tokens) and vary by less than 0.5 points over the remaining 0.7 epochs.
The average text score stays within 0.9 points of the model without mid-training across all evaluated training steps.

Figure~\ref{fig:progress-all} breaks down individual benchmark trajectories.
After full mid-training, the largest video gains are on long-form egocentric understanding (EgoSchema) and general video question answering (NExT-QA, VideoMME), while fine-grained temporal reasoning (TempCompass) improves by 1.5 points.
Averaged by task type, the largest image gains at the end of mid-training are on text-rich image understanding (+5.8 points) and general perception and diagram understanding (+5.2 points), followed by visual reasoning (+4.4 points) and text recognition (+3.2 points).
Across all evaluation intervals, scores on commonsense, knowledge, coding, and GPQA remain within 3 points of the baseline.
In contrast, reasoning benchmarks (GSM8K, MATH500, BBH) fluctuate by up to 9 points.

\subsection{Visual Next-Token Prediction Is Sufficient Without Annotations}
\label{subsec:objective}

Table~\ref{tab:ablation} demonstrates that purely self-supervised visual next-token prediction is sufficient for effective mid-training, matching or outperforming objectives that supervise with frame-level dense captions.
We compare our approach against two caption-based targets, namely current-caption and next-caption prediction, where the continuous visual target is replaced by a text caption of the current or next frame (see Section~\ref{subsubsec:captioning}).

Predicting next-frame captions yields no empirical advantage over predicting visual tokens directly. 
It yields similar video and image scores, but degrades language performance by 1.5 points. 
Even when controlling for training budget by comparing against our 0.5-epoch visual model (52.37 video, 53.76 image, 48.32 text), visual next-token prediction matches vision gains within 0.3 points while better preserving text capability. 
Current-caption prediction likewise fails to offer a compelling alternative, falling 5.0 points behind visual tokens on image understanding and showing inconsistent transfer across tasks.
These comparisons establish that raw visual token prediction is sufficient on its own, allowing video mid-training to remain fully self-supervised while avoiding the computational overhead and label noise of caption generation.

\begin{table}[t]
\caption{Comparison of mid-training objectives on YT-1B.
All models are instruction-tuned on LLaVA-OneVision-Data, and No mid-training is the same model instruction-tuned without mid-training.
The best value in each column is in bold.
}
\vspace{0.1cm}
\label{tab:ablation}
\centering
\begin{tabular}{lccc}
\toprule
Objective & Video & Image & Text \\
\midrule
No mid-training      & 49.67 & 48.72 & 48.03 \\
\midrule
Current caption     & 51.28 & 48.77 & 47.44 \\
Next caption        & 52.51 & 53.54 & 47.40 \\
Visual next token   & \textbf{52.59} & \textbf{53.80} & \textbf{48.89} \\
\bottomrule
\end{tabular}
\vspace{-0.3cm}
\end{table}


\section{Discussion}
\label{sec:discussion}

We propose autoregressive next-token prediction on raw web video as a mid-training objective without text supervision that bridges language pretraining and multimodal instruction tuning.
Despite seeing only continuous video frames, the model gains 2.9 points on video-language and 5.1 points on image-language benchmarks, with the greatest improvements in text-rich image understanding and general perception.
Predicting next-frame captions offers no advantage over predicting raw visual tokens, demonstrating that video mid-training can remain entirely self-supervised without the expense or label noise of captioning pipelines.

Although mid-training has no text loss, the mid-trained model averages 48.9 on text benchmarks after instruction tuning, compared with 48.0 for the model without mid-training.
Both models score below the original language model on GSM8K, MATH500, and BBH, and for the model without mid-training this drop comes from instruction tuning alone.
While the mid-trained model degrades less, its 5.5-point margin on BBH is smaller than the 11.7-point discrepancy between instruction-tuning runs from the same initial state.

Downstream multimodal gains appear within the first 0.3 epochs, after which scores change by less than 0.5 points across the remainder of training. 
Whether denser temporal sampling or longer sequence horizons can yield further improvements remains an open question. 
In addition, determining whether these gains stem from temporal structure or simply from additional exposure to visual tokens requires future frame-shuffling experiments.
Evaluating these dynamics across larger language models will establish whether uncaptioned video mid-training can serve as a standard precursor to multimodal instruction tuning.

\clearpage
\newpage
\bibliographystyle{assets/plainnat}
\bibliography{references}


\end{document}